\pdfoutput=1

\documentclass[11pt]{article}
\usepackage{graphicx}
\usepackage{booktabs,tcolorbox,epsfig,balance}
\usepackage{subfig}

\usepackage{multirow}
\usepackage{multicol}
\usepackage{graphicx}

\usepackage[]{emnlp2021}

\usepackage{times}
\usepackage{latexsym}

\usepackage[T1]{fontenc}

\usepackage[utf8]{inputenc}

\usepackage{microtype}

\title{Ensemble Fine-tuned mBERT for Translation Quality Estimation}

\author{Shaika Chowdhury \thanks{work done during internship at IQVIA} \\
  University of Illinois at Chicago, US \\
  \texttt{schowd21@uic.edu} \\\And
  Naouel Baili \\
  IQVIA, US \\
  \texttt{naouel.baili@iqvia.com} \\ \AND
  Brian Vannah \\
  IQVIA, US \\
  \texttt{brian.vannah@iqvia.com} \\}

\begin{document}
\maketitle
\begin{abstract}
Quality Estimation (QE) is an important component of the machine translation workflow as it assesses the quality of the translated output without consulting reference translations. In this paper, we discuss our submission to the WMT 2021 QE Shared Task. We participate in Task 2 sentence-level sub-task that challenge participants to predict the HTER score for sentence-level post-editing effort. Our proposed system is an ensemble of multilingual BERT (mBERT)-based regression models, which are generated by fine-tuning on different input settings. It demonstrates comparable performance with respect to the Pearson's correlation and beats the baseline system in MAE/ RMSE for several language pairs. In addition, we adapt our system for the zero-shot setting by exploiting target language-relevant language pairs and pseudo-reference translations.
\end{abstract}








\section{Introduction}
Progress in machine translation (MT) has accelerated due to the introduction of deep learning based approaches, dubbed as neural machine translation (NMT) \cite{cho2014learning,sutskever2014sequence,bahdanau2014neural}. Several metrics (e.g., BLEU \cite{papineni2002bleu}, METEOR \cite{agarwal2008meteor}) are used to automatically evaluate the quality of the translations outputted by the NMT systems. However, these evaluation metrics require comparing the NMT outputs against human-prepared reference translations, which cannot be readily obtained. To tackle this predicament,  recently quality estimation (QE) \cite{blatz2004confidence,specia2018quality} has emerged as an alternative evaluation approach for NMT systems. QE obviates the need for human judgements and hence can be efficiently integrated into the dynamic translation pipeline in the industry setting.  

QE is performed at different granularity (e.g., word, sentence, document) \cite{kepler2019unbabel}; in this work we focus on the sentence-level post-editing effort task, which predicts the quality of the translated sentence as a whole in terms of the number of edit operations that need to be made to yield a post-edited translation, termed as HTER \cite{snover2006study}.

Sentence-level QE using neural approaches is generally treated as a supervised regression problem involving mainly two steps. In the first step, an encoder is used to learn vector representation/s of the source and translation sentences. While in the second step, the learned representations are passed through a sigmoid output layer to estimate the HTER score. These two steps can be performed either with a single model in an end-to-end fashion (e.g., Bi-RNN \cite{ive2018deepquest}), or using two separate models (e.g., POSTECH \cite{kim2017predictor}). The different QE systems vary in their choice of the encoder, which range from RNN-based to Transformer-based models.

In this work, we leverage the fine-tuning capability of a Transformer-based encoder, namely the mBERT \cite{devlin2018bert} pre-trained model. Alongside the standard practice of feeding both the source and target (i.e., translation) sentences as the input sequence \cite{kepler2019unbabel, kim2019qe}, we also explore other input settings based on only the target-side sentences (i.e., monolingual context). To this end, our final QE system is an ensemble of several mBERT models \footnote{we also experimented with XLM-RoBERTa \cite{conneau2019unsupervised} as the component model in our preliminary run; however, the results were worse compared to mBERT}, each generated by fine-tuning on a different input combination comprising the source and/or target sentences. We experiment with the following three input settings: (1) both source and target, (2) just target and (3) both target and a randomly-sampled target sentence in the data forming the input sequence. Empirical analysis on 6 language
pairs shows that the ensemble model is able to perform better than the individual
fine-tuned models.
Moreover, we provide experimental results for zero-shot QE, where training data for the test language pair is not available. This we tackle by improvising on the available training/dev data that match the target language of the test language pair and also by generating the pseudo-reference translations in that language.



\section{Data}
We use the WMT21 QE Shared Task 2 sentence-level data \cite{fomicheva2020mlqepe,tacl2020} for the following 7 language pairs: English-German (En-De), Romanian-English (Ro-En), Estonian-English (Et-En), Nepalese-English (Ne-En), Sinhala-English (Si-En),  Russian-English (Ru-En) and Khmer-English (Km-En). Source-side data for each language pair includes sentences from Wikipedia articles, with part of the data gathered from Reddit articles for Ru-En. To obtain the translations, state-of-the-art MT models \cite{vaswani2017attention} built using fairseq toolkit \cite{ott2019fairseq} were used. The label for this task is the HTER score for the source-translation pair. Annotation was performed first at the word-level with the help of TER \footnote{http://www.cs.umd.edu/~snover/tercom/} tool. The word-level  tags were then aggregated deterministically to obtain the sentence-level HTER score. The training, development, test and blind test data sizes for each language pair (except Km-En) are 7K, 1K, 1K and 1K instances respectively. As Km-En language pair was introduced for zero-shot prediction, only the test data containing 990 source and translation sentences was provided.  

\section{Our Approach}
A key innovation in recent neural models lies in learning
the contextualized representations by pre-training on a language modeling task. One such model, the multilingual BERT (mBERT) \footnote{https://github.com/google-research/bert/blob/master/multilingual.md}, is a transformer-based masked language model that is pre-trained on monolingual Wikipedia corpora of 104 languages with a shared word-piece vocabulary. Training the pre-trained mBERT model for a supervised downstream task, aka \textit{finetuning}, has dominated performance across a wide spectrum of NLP tasks \cite{devlin2018bert}. Our proposed approach leverages this fine-tuning capability of mBERT so as to form the component models in the ensemble QE system (Section 3.3). That is, each component model is a re-purposed mBERT that is fine-tuned for the sentence-level HTER score prediction task on one of the three input settings discussed in Section 3.2.

\subsection{Fine-tuning mBERT for Regression}
mBERT's model architecture is similar to BERT \footnote{https://huggingface.co/bert-base-uncased} and contains the following parameter settings: 12 layers, 12 attention heads and 768 hidden dimension per token. However, the only difference is that mBERT is trained on corpora of multiple languages instead of just on English. This enables mBERT to share  representations across the different languages and hence can be conveniently used for all language pairs in the WMT21 data. 

We first load the pre-trained mBERT model \footnote{https://huggingface.co/bert-base-multilingual-uncased} and use its weights as the starting point of fine-tuning. The pre-trained mBERT is then trained on  QE-specific input sequences (Section 3.2) for a few epochs such that the constructed sequence \texttt{$X$} is consumed by mBERT to output the contextualized representation $\textbf{h} = (h_{CLS},h_{x_1},h_{x_2},..h_{x_T}, h_{SEP})$. Here, $[CLS]$ is a special symbol that denotes downstream classification and $[SEP]$ is for separating non-consecutive token sequences. Considering the final hidden  vector of the $[CLS]$ token as the aggregate representation, it is then passed into the output layer with sigmoid activation to predict the HTER score:


    
    \begin{equation}
        \large
    y = sigmoid(\mathbf{W}\cdot \mathbf{h}_{CLS} + \mathbf{b})
    \end{equation}
    
    $\textbf{W}$ is a weight matrix for sentence-level QE fine-tuning that is trained along with all the parameters of mBERT end-to-end.

\subsection{Input Settings}
We construct the input sequence for each language pair in the following three ways: 
\begin{enumerate}
    \item[]\textbf{SRC-MT:}
    Given a source sentence $\textbf{s} = (s_1,s_2,...s_N)$ from a source language (e.g., English) and its translation $\textbf{t} = (t_1,t_2,...t_M)$ from a target language (e.g., German), we concatenate them together as $ X =  ([CLS],t_1,t_2,...t_M, [SEP],  s_1,s_2,...s_N,   \\\ [SEP])$ to form the input sequence.  
    
    \item[]\textbf{MT:}
    The target sentence is only used to form the input sequence, $ X = ([CLS],t_1,t_2,...t_M, [SEP])$.
    

    \item[]\textbf{MT-MT':}
    
    Given the translation 
    $\textbf{t}$ for a source sentence $\textbf{s}$, we randomly sample another translation $\textbf{t'} = (t'_1,t'_2,...t'_K)$ from the training data having HTER label close to $\textbf{t}$ \footnote{to ensure that $\textbf{t'}$ is similar to \textbf{t}, we check that the difference between their HTER scores is within 0.1}. Although the source sentences for $\textbf{t}$ and $\textbf{t'}$ are different, we assume the additional monolingual context would help mBERT learn the correlating QE-specific features between $\textbf{t}$ and $\textbf{t'}$ for the target-side language. The resultant input sequence is $ X =  ([CLS],t_1,t_2,...t_M, [SEP],  t'_1,t'_2,...t'_K,   \\\ [SEP])$.

We fine-tune each of these mBERT models using AdamW optimizer \cite{kingma2014adam,loshchilov2017decoupled} for two epochs with a batch size of 32 and a learning rate of $2e^{-5}$.



\end{enumerate}
\subsection{Ensemble Model}
To take advantage of the individual strengths of the three mBERT component models fine-tuned on the aforementioned input settings, we combine their HTER score predictions by training an ensemble model. In particular, we experiment with three different ensemble models - Gradient Boosting  \cite{friedman2001greedy}, AdaBoost \cite{freund1997decision} and Average. For Gradient Boosting and AdaBoost we use the implementation in scikit-learn \footnote{https://scikit-learn.org} with 10-fold cross validation. The settings for Gradient Boosting are: number of estimators 600, learning rate 0.01, minimum number of samples 3 and other default settings. We use the default settings for AdaBoost. In Average ensembling, we average the HTER score predictions by the three mBERT models. Our system submission to WMT21 is based on Gradient Boosting as it gave the best performance on the test data, as shown in Table \ref{tab::Table0}.

    \begin{table}[h!]
\caption{Performance of ENSBRT with different ensemble methods on the En-De test set.}
\label{tab::Table0}
\begin{tabular}{l|lll}
\toprule 
                   {}& \multicolumn{3}{l}{} \\
             & \textbf{Avg}   & \textbf{AdaBoost} & \textbf{GradBoost}     \\ \midrule

Pearson's     & 0.266   &0.458 & \textbf{0.473}

\\
Spearman's     & 0.249   &0.436 & \textbf{0.443}

\\\bottomrule

\end{tabular}
\end{table}

    \begin{table*}[h!]
\centering
\caption{Performance in Pearson's correlation of mBERT fine-tuned with different input settings on the test set. ENSBRT is the proposed ensemble mBERT QE system.}
\label{tab::Table1}
\begin{tabular}{l|llllll}
\toprule 
                   {}& \multicolumn{6}{l}{} \\
             & \textbf{En-De}   & \textbf{Ro-En} & \textbf{Ru-En} &\textbf{Si-En}   & \textbf{Et-En} &\textbf{Ne-En}    \\ \midrule

SRC-MT    &0.389     &0.793 &0.400  &0.526 &0.601 & 0.489 

\\
MT   &0.469     &0.762 &0.374  &0.552 &0.580 & 0.491
\\

MT-MT'  & 0.431    &0.761 &0.350  &0.492 &0.556 & 0.454

\\

ENSBRT & \textbf{0.473} &\textbf{0.802} &\textbf{0.418}  &\textbf{0.576} &\textbf{0.632} & \textbf{0.525}
  
\\\bottomrule

\end{tabular}
\end{table*}

    \begin{table*}[h!]
\centering
\caption{Performance of BASELINE and ENSBRT on the WMT21 blind test set for different language pairs. Bold indicates ENSBRT beats BASELINE in that metric.}
\label{tab::Table2}
\begin{tabular}{l|lllllll}
\toprule 
                   {}& \multicolumn{7}{l}{} \\
             & \textbf{En-De}   & \textbf{Ro-En} & \textbf{Ru-En} &\textbf{Si-En }  & \textbf{Et-En} &\textbf{Ne-En} &\textbf{Km-En}  \\ \midrule

\multirow{3}{*}{\rotatebox[origin=c]{90}{\textit{BASELINE}}} \hspace{0.17cm}Pearson's $\uparrow$  &0.529     &0.831 &0.448 &0.607 &0.714 &0.626 &0.576

\\
\hspace{0.42cm} MAE $\downarrow$   &0.183    &0.142 &0.255 &0.204 &0.195 &0.205 &0.241
\\
\hspace{0.42cm} RMSE $\downarrow$   &0.129     &0.115 &0.188 &0.159 &0.149 &0.160 &0.196

\vspace{0.4cm}

\\ \midrule

\multirow{3}{*}{\rotatebox[origin=c]{90}{\textit{ENSBRT}}} \hspace{0.17cm}Pearson's $\uparrow$  &0.519     &0.795 &0.376 &0.522 &0.666 &0.572 &0.529

\\
\hspace{0.42cm} MAE $\downarrow$   &\textbf{0.171}    &0.171 &\textbf{0.251} &0.206 &\textbf{0.171} &\textbf{0.176} &0.262
\\
\hspace{0.42cm} RMSE $\downarrow$   &0.129     &0.141 &0.189 &0.162 &\textbf{0.132} &\textbf{0.139} &0.197
  
\\\bottomrule

\end{tabular}
\end{table*}

\subsection{Zero-Shot QE}

Performing sentence-level QE in the zero-shot setting presents a unique challenge as the QE system is expected to predict HTER scores for sentences in a test language pair (e.g., Km-En) without having been trained on any instances from that test language pair. We address this by training on language pairs in the WMT21 QE data that match the target-side language (i.e., En) in the test language pair. The reason we focus on the target-side language is because the component mBERT models in the proposed ensemble QE system are fine-tuned on monolingual input sequences in the target-side language, which could potentially help the QE system generalize on the unseen test language pair. We consider the training and development data for the following language pairs in WMT21 QE data: Ro-En, Si-En, Et-En. Additionally, we augment this data by generating pseudo-references in the target language. A \textit{pseudo-reference} \cite{scarton2014document} is a translation for a source sentence that is outputted by a different NMT system than the one that produced the actual translations (e.g., transformer-based translation system proposed in \cite{vaswani2017attention}) and has shown to improve sentence-level QE performance \cite{soricut2012combining}. We use Google Translate \footnote{https://github.com/lushan88a/google\_trans\_new} to get the pseudo-references in En for the Ro, Si and Et source sentences. The HTER scores for the translation and pseudo-reference pairs are then obtained using the TER tool. We train the ensemble QE system on the combined WMT21 QE data and the pseudo-reference parallel data, and test on the unseen test language pair. 

\section{Baseline}
The baseline QE system (BASELINE) set by the WMT21 organizers this year is the Transformer-based Predictor-Estimator model \cite{kepler2019openkiwi,moura2020unbabel}. XLM-RoBERTa is used as the Predictor for feature generation. The baseline system is fine-tuned on the HTER scores and word-level tags jointly.  

\section{Results}
  
\begin{figure*}
\begin{multicols}{2}
    \includegraphics[width=1.0\linewidth]{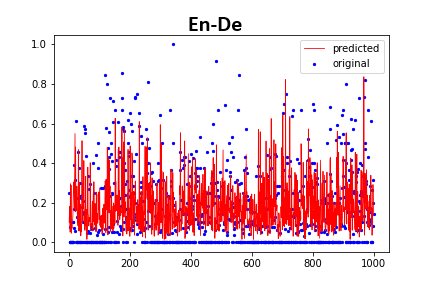}\par
   \includegraphics[width=1.0\linewidth]{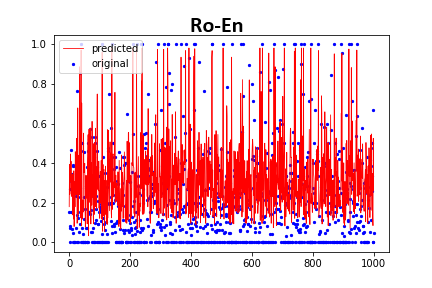}\par 
    \end{multicols}
\begin{multicols}{2}
    \includegraphics[width=1.0\linewidth]{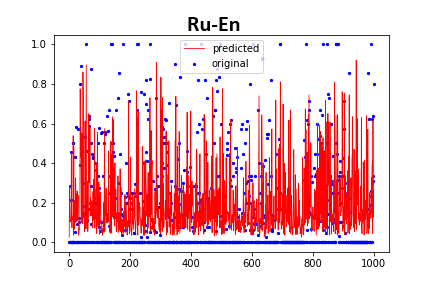}\par
    \includegraphics[width=1.0\linewidth]{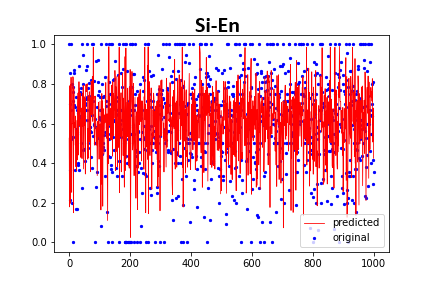}\par
\end{multicols}
\begin{multicols}{2}
    \includegraphics[width=1.0\linewidth]{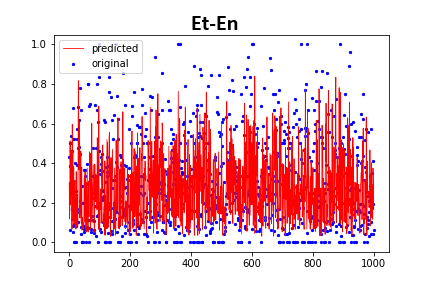}\par
    \includegraphics[width=1.0\linewidth]{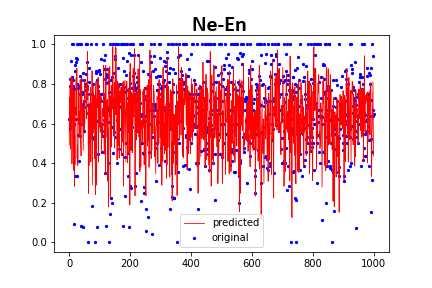}\par
\end{multicols}
\caption{Visualization comparing HTER score predictions by ENSBRT (i.e., predicted (red)) against the gold labels (i.e., original (blue)) for 6 language pairs on the test set. X-axis represents each data point and Y-axis is the HTER score. The closer the corresponding red line and blue dot are to each other the better, as we expect the HTER prediction to be closer to the ground truth.}
\label{visall}
\end{figure*}

 Table \ref{tab::Table1} 
 presents the experimental results of mBERT fine-tuned on the $SRC$-$MT$, $MT$ and $MT$-$MT'$ input settings, as well as the performance of the  ensemble of the three mBERT models, which we call \textit{ENSBRT}. First, comparing among the three input settings,  it seems that mBERT exhibits competitive results even when it does not have knowledge of the source-side text in the $MT$ and $MT$-$MT'$ settings, in particular for the following language pairs - En-DE, Si-En, Ne-En. While the ensemble mBERT model, ENSBRT, outperforms the independent counterparts for all the language pairs. This shows that the ensemble method can help to balance out the weakness of any component model,  thereby benefiting the sentence-level QE task overall. We also visualize ENSBRT's predictions against the ground truth HTER scores in Figure \ref{visall}.

Table \ref{tab::Table2} compares the QE performance between the BASELINE and ENSBRT in terms of Pearson's correlation, RMSE and MAE on the WMT21 blind test set, for which the ground truth HTER scores were not available at the time. We submitted results for 6 language pairs (En-De, Ro-En, Ru-En, Si-En, Et-En, Ne-En) in the normal QE setting and one language pair (Km-En) for zero-shot prediction. ENSBRT demonstrates comparable performance to the BASELINE for Pearson's and outperforms it in either MAE or RMSE for the following language pairs: En-De, Ru-En, Et-En and Ne-En.

\section{Conclusion}
In this work, we describe the \textit{ENSBRT} system submission to the WMT21 QE Shared Task. ENSBRT is based on fine-tuning the multilingual BERT pre-trained model for sentence-level translation quality score prediction. We explore three different input settings for fine-tuning which include either bilingual or monolingual context, and combine the predictions of the three models using ensemble methods as our final system. Furthermore, zero-shot QE is facilitated by using labeled data for existing language pairs and pseudo-references that align with the target language of the unseen test data.

\bibliography{anthology,custom}
\bibliographystyle{acl_natbib}




\end{document}